\documentclass{IEEEmce}

\usepackage[colorlinks,urlcolor=blue,linkcolor=blue,citecolor=blue]{hyperref}
\usepackage{amsmath, amssymb} 
\usepackage{booktabs}   
\usepackage{hyperref}
\hypersetup{ 
 colorlinks=true, 
 linkcolor=red, 
 filecolor=blue, 
 citecolor = blue, 
 urlcolor=blue, 
 } 

\usepackage{soul}
\sethlcolor{yellow}

\usepackage{upmath}

\jvol{XX}
\jnum{XX}
\paper{XX}
\jmonth{xxx/xxx}
\publisheddate{DD MM YYYY}
\currentdate{DD MM YYYY}
\jname{IEEE Consumer Electronics Magazine}
\pubyear{YYYY}
\doiinfo{MCE.YYYY.Doi Number}

\begin{document}

\sptitle{Feature Article: Collaborative Intelligence} 


\title{Feature Coding for Scalable Machine Vision}


\author{Md Eimran Hossain Eimon, Juan Merlos, Ashan Perera, Hari Kalva, Velibor Adzic, and Borko Furht}
\affil{Florida Atlantic University}



\markboth{Collaborative Intelligence}{Feature Coding for Scalable Machine Vision}


\begin{abstract} Deep neural networks (DNNs) drive modern machine vision but are challenging to deploy on edge devices due to high compute demands. Traditional approaches—running the full model on-device or offloading to the cloud—face trade-offs in latency, bandwidth, and privacy. Splitting the inference workload between the edge and the cloud offers a balanced solution, but transmitting intermediate features to enable such splitting introduces new bandwidth challenges. To address this, the Moving Picture Experts Group (MPEG) initiated the Feature Coding for Machines (FCM) standard, establishing a bitstream syntax and codec pipeline tailored for compressing intermediate features. This paper presents the design and performance of the Feature Coding Test Model (FCTM), showing significant bitrate reductions—averaging 85.14\%—across multiple vision tasks while preserving accuracy. FCM offers a scalable path for efficient and interoperable deployment of intelligent features in bandwidth-limited and privacy-sensitive consumer applications. \end{abstract}

\maketitle

\enlargethispage{10pt}


\begin{figure*}
    \centering
    \includegraphics[width=1\linewidth]{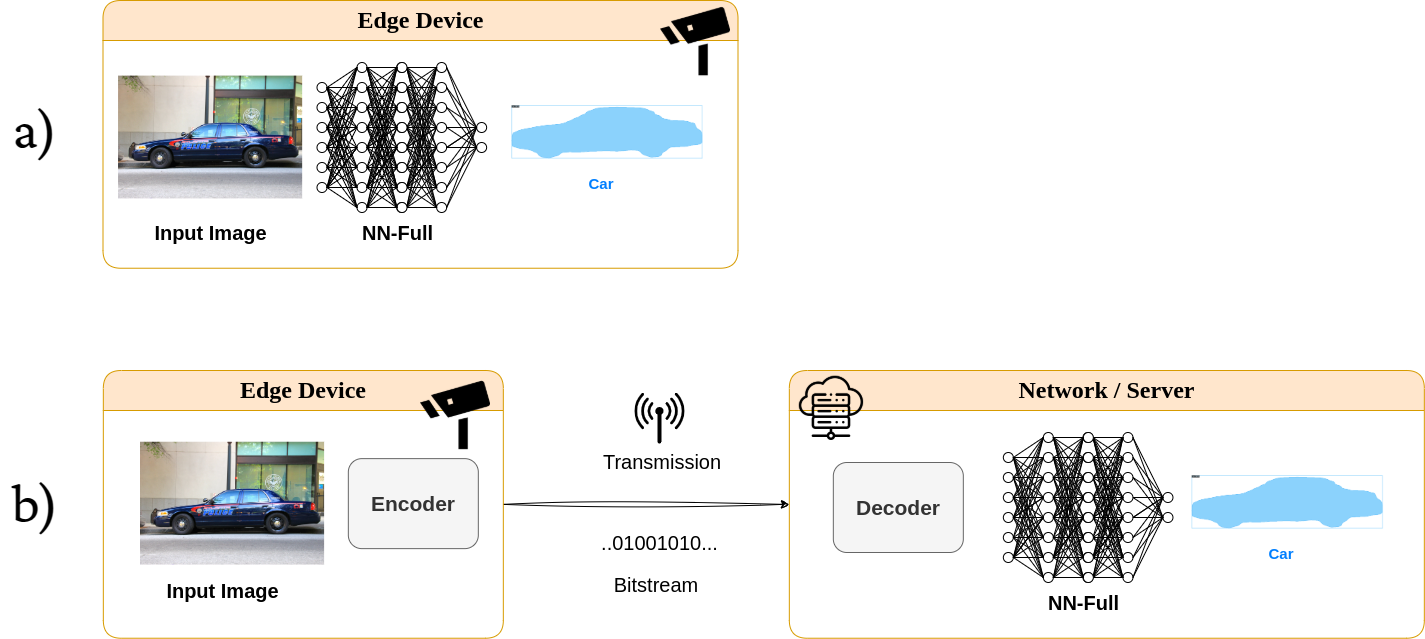}
    \caption{a) Edge inference: all neural network computation is done on edge device, b) Remote inference: all neural network computation is done on remote server}
    \label{fig:rem}
\end{figure*}
\begin{figure*}[t]
    \centering
    \includegraphics[width=1\linewidth]{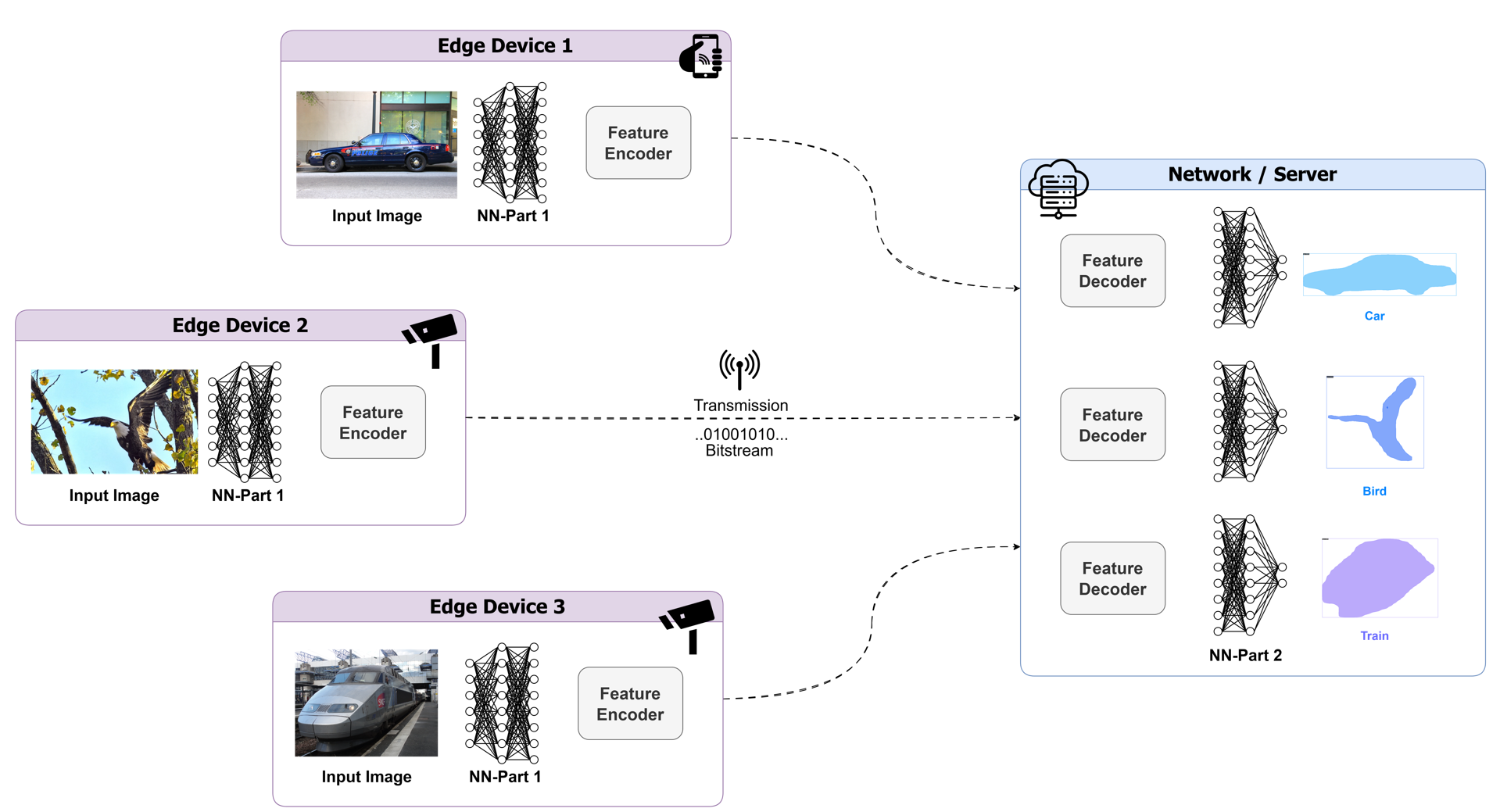}
    \caption{Split inferencing architecture where part one of a neural network (NN-Part 1) is deployed on the edge device while part two of the neural network (NN-Part 2) is deployed on the cloud.}
    \label{fig:intro}
\end{figure*}

\chapterinitial{Modern computer vision systems} are integral to an expanding range of intelligent consumer applications. Tasks such as object detection, semantic segmentation, and object tracking enable smart features in mobile phones, autonomous vehicles, augmented reality (AR) headsets, and home surveillance devices. These functions are typically powered by deep neural networks (DNNs) which require high computation to process visual data in real time. As the complexity of these models grow, deploying them efficiently on power and resource constrained edge devices has become a fundamental challenge. To alleviate device-side burden, systems are traditionally designed to rely on a remote inference model wherein image or video data is compressed and transmitted from the edge device to a remote server/cloud platform for full inference. This approach allows cloud infrastructure to host powerful models and respond with high-accuracy predictions if needed. However, this approach is resource hungry and poses scalability challenges.

Figure~\ref{fig:rem} shows the edge and remote inference use cases with the computational complexity either largely on the edge device or the cloud. As the number of deployed edge devices scale up, the burden of performing the complete inference shifts entirely to the cloud, leading to network congestion, high server costs, and increased response latency. Meanwhile, the local computing capacity of edge devices, although often insufficient for full inference, remains underutilized. As consumer electronics evolve to include more on-device AI accelerators and neural processing units (NPUs), there is a growing incentive to redesign AI pipelines to share the computational load more effectively between the device and the cloud.

Split inference~\cite{icce}, also known as collaborative intelligence~\cite{colabintel}, offers a promising solution. In split inference, as shown in Figure~\ref{fig:intro}, a neural network is partitioned such that early layers are executed on the edge device, generating intermediate features. These features are then transmitted to the cloud, where the remaining computation is performed. This architecture leverages the device’s available compute and enables more efficient load balancing between clients and servers. This solution also offers strong privacy since the intermittent network features are transmitted rather than the actual video~\cite{split_inf}. However, transmitting intermediate features introduces its own challenge: depending on the neural network’s split point, these tensors are often high-dimensional and data-intensive, potentially requiring more bandwidth than that needed for streaming the corresponding video~\cite{rate_dis}.

Recognizing this importance, the Moving Picture Experts Group (MPEG), operating under ISO/IEC JTC1/SC29/WG04, initiated the \textbf{Feature Coding for Machines (FCM)} standardization effort to develop efficient technologies to support split inference. This activity focuses on defining standardized bitstreams and compression tools specifically for neural intermediate features in machine vision systems. Unlike traditional video codecs optimized for human perception, FCM is tailored for downstream AI tasks even at reduced bitrate. In response to MPEG’s formal Call for Proposals (CfP) in 2023~\cite{press_release_mpeg144}, a diverse set of proponents answered the call, including Canon, InterDigital, ETRI, Sharp, Digital Insights Inc., China Telecom, ZJU, KHU, HNU, KAU, and Florida Atlantic University (FAU). Out of this collaborative process emerged the Feature Coding for Machines (FCM) standard. 

Recent studies have further advanced split inference for machine vision. Zhang et al.~\cite{afc} proposed an asymmetrical feature coding strategy designed to optimize coding efficiency and task performance across heterogeneous vision tasks by assigning task‑specific feature representations. Complementing this, Liu et al.~\cite{mf_bit_alloc} developed a multiscale feature importance prediction framework that dynamically allocates bits among feature scales based on their task relevance.


While FCM employs an autoencoder-style feature reduction and restoration framework, the standard's role and objectives are fundamentally different from an autoencoder. Autoencoders are closed systems that learn a latent representation optimized for reconstructing the input signal, typically by minimizing pixel-level distortion~\cite{nokia_vcm}. Such representations are bound to a specific encoder–decoder pair, limiting interoperability across devices or networks. By contrast, FCM is explicitly designed as a standardized feature coding framework: it defines a common bitstream syntax, side-information signaling (e.g., global statistics), and packing strategies so that any compliant decoder can interpret features generated by any compliant encoder. This standardization enables collaborative intelligence, where heterogeneous devices and cloud servers can exchange semantically rich features rather than raw video. In large-scale smart city deployments, for example, thousands of cameras and sensors from different manufacturers must interoperate seamlessly to support applications such as traffic management, crowd monitoring, and public safety. FCM's standardized intermediate feature representations allow these devices to transmit compact, task-relevant features that can be aggregated and analyzed jointly across the network, ensuring low latency, reduced bandwidth, and privacy preservation. This interoperability and scalability position FCM as a strong foundation for advancing next-generation, distributed scalable machine vision systems.

This article first introduces representative real-world deployment scenarios that motivate the need for efficient feature compression. It provides a technical overview of MPEG’s reference implementation—the Feature Coding Test Model (FCTM), and then summarizes experimental results based on the standardized benchmark defined by industry experts to evaluate feature compression performance called the Common Test and Training Conditions (CTTC)~\cite{fcm_cttc}. Finally, the article analyzes trade-offs between task accuracy, compression efficiency, and computational complexity; then it concludes by identifying current limitations and outlining future research directions to enhance FCM’s scalability, interoperability, and suitability for real-time, low-power applications.

\section{REAL-WORLD USE CASES}

\begin{figure}[t]
    \centering
    \begin{minipage}[b]{1\linewidth}
    \centering
    \includegraphics[width=1\textwidth]{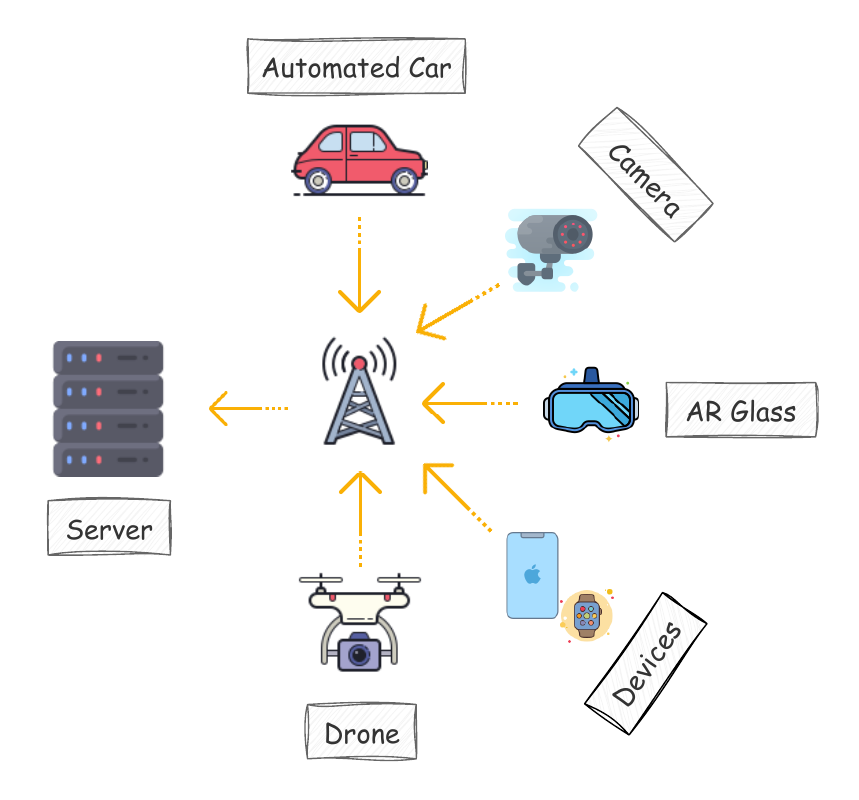}
    \end{minipage}
\caption{Real-world deployment scenarios of FCM, where edge devices execute early network layers and transmit intermediate feature maps to cloud servers for completing inference with the remaining layers.}

\vspace{-0.5cm}
\label{fig:use_cases}
\end{figure}

FCM enables scalable, low-latency, and privacy-aware intelligence across diverse consumer applications through efficient, interoperable feature compression. Some of the most impactful use cases—ranging from smart surveillance to connected vehicles and AR wearables—are illustrated in Figure~\ref{fig:use_cases} and discussed in detail below.

\subsection{Smart Surveillance}

In modern surveillance systems—deployed in retail stores, public spaces, or industrial kitchens—real-time analytics are crucial for detecting security threats, safety violations, or behavioral anomalies. FCM allows edge devices to transmit compressed, task-relevant features rather than full-resolution video streams, significantly reducing bandwidth usage and preserving privacy. For instance, in an intelligent kitchen, action recognition and outfit compliance can be performed efficiently without streaming raw video to the cloud.

\subsection{Cooperative and Connected Vehicles}

Vehicles in intelligent transportation networks often need to share sensory insights to support collision avoidance, traffic prediction, and situational awareness. FCM makes it possible to compress and transmit features from in-vehicle sensors—such as dashcams—through communication links. This enables vehicles and roadside infrastructure to collaborate in real time. Standardized FCM bitstreams ensure interoperability across manufacturers, enhancing the safety and efficiency of connected mobility systems.

\subsection{AR Glasses and Smart Wearables}

Wearable AI devices like augmented reality glasses are constrained by power, compute, and connectivity limitations. FCM supports these platforms by allowing them to offload computation through the transmission of compressed intermediate features, enabling the execution of state-of-the-art deep learning models while reducing power consumption.

\subsection{Multi-Camera Retail Analytics}

Smart retail environments use multiple cameras to monitor customer behavior, track inventory, and detect security incidents. FCM allows these cameras to share semantically rich but compact features between devices or with central servers. This reduces transmission overhead and enables faster, more coordinated analytics for customer experience optimization, loss prevention, and dynamic inventory tracking.

\subsection{Scalable Smart City Infrastructure}

Urban environments depend on large-scale sensor networks for real-time decision-making in traffic control, crowd management, and public safety. As the number of deployed devices grows, transmitting raw visual data becomes unsustainable. FCM provides a scalable alternative by enabling these devices to send compressed, standardized feature representations optimized for downstream machine vision tasks. This makes FCM a key enabler for smart city systems that demand real-time, interoperable, and privacy-aware machine vision.

\begin{figure*}[t]
    \centering
    \begin{minipage}[b]{1\linewidth}
    \centering
    \includegraphics[width=1\textwidth]{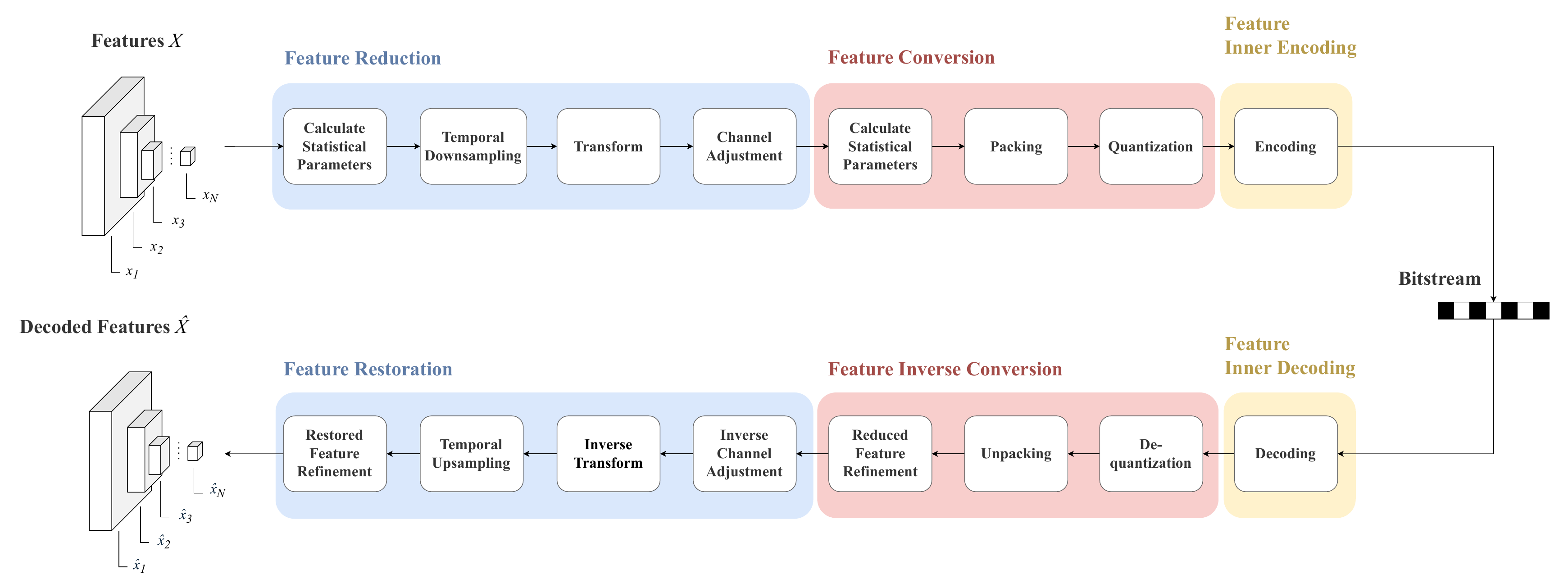}
    \end{minipage}
\caption{Overview of Feature Coding for Machines. The encoder takes features $X$ to produce a bitstream. Afterwards, the decoder takes a bitstream to produce features $\hat{X}$.}
\vspace{-0.5cm}
\label{fig:overview}
\end{figure*}
\section{OVERVIEW OF FCM}
The Feature Coding Test Model (FCTM)~\cite{fctm} is the implementation of FCM used in the standardization process. FCTM compresses a time-series of intermediate features that each may have a different spatial and channel resolution, as is the assumption when extracting from an arbitrarily split neural network. Formally, at a given point in time, a set of intermediate features \(X\) may be represented as \(X=\{x_n\}^{N}_{n=1}\), where \(x_n\in \mathbb{R}^{C_n \times H_n \times W_n}\) is an individual feature layer and N denotes the total number of intermediate feature layers. Figure~\ref{fig:overview} illustrates the compression and decompression of a series of \(X\) features to produce a series of semantically similar \(\hat{X}\) features at a lower bitrate than alternative codecs. The encoder has three stages: Feature Reduction, Feature Conversion, and Feature Inner Encoding. Feature Reduction transforms the features into a latent space with reduced entropy spatially and temporally. Feature Conversion adapts the reduced features from latent space into a commonly operable video structure, where sets of feature layers are frames arranged in receiving order. Feature Inner Encoding compresses the video data with a traditional video codec, such as H.26x compliant codecs~\cite{avc, hevc, vvc}, while incorporating additional feature-level information into the bitstream. The decoder follows a series of corresponding inverse stages: Feature Inner Decoding, Feature Inverse Conversion, and Feature Restoration. In essence, FCTM is an adapter from features to video with the following key advantages: 1) minimal feature entropy, and 2) support for the decades-long development of modern video compression. The following sections provide a brief description of each tool in the FCTM encoder and decoder.

\section{FEATURE REDUCTION}

The Feature Reduction module contains the following four tools: 1) Calculate Statistical Parameters, 2) Temporal Downsampling, 3) Transform, and 4) Channel Adjustment.

\subsection{Calculate Statistical Parameters}
Vision inference accuracy can be largely preserved provided that the decoded intermediate features retain statistical properties closely aligned with those of the original features in the latent space~\cite{ce42a, ce42c}. To mitigate potential distributional shifts introduced by quantization, statistical parameters of the intermediate feature tensor $X$ are computed at the encoder and transmitted to the decoder. Assuming that the $N$ intermediate feature layers $\{x_{1}, x_{2}, \dots, x_{N}\}$ are statistically independent and normally distributed, the global mean $\mu_{X}$ and standard deviation $\sigma_{X}$ are calculated as~\cite{eimran_iscas}:

\begin{align}
\mu_{X} &= \sum_{n=1}^{N} \mu(x_{n}) \\
\sigma_{X} &= \sqrt{\sum_{n=1}^{N} \sigma^2(x_{n})}
\end{align}

To reduce signaling overhead, these parameters are quantized into 16-bit \textit{bfloat} (Brain Floating Point) representations in place of conventional 32-bit floating point encoding~\cite{ce42c_simplified} and periodically transmitted.

\subsection{Temporal Downsampling}

Temporal redundancies arise when there is minimal motion between sets of feature layers. In such cases, transmitting every single set of feature layers becomes inefficient and unnecessary~\cite{m67615}. To address this, the encoder can temporally sample at regular intervals. For example, a sampling ratio of $2\times$ can be used, meaning every other set of layers is discarded, effectively halving the total count and the amount of information necessary in the bitstream. For an odd number of sets, the last set must be preserved. The decoder will be responsible for reconstructing discarded layers from preserved layers. Under the assumption of minimal motion, reconstructed layers are expected to have little discrepancy. Increases in sampling ratio decreases the bitrate, although only a $1\times$ or $2\times$ sampling ratio is supported. A $1\times$ sampling ratio effectively leaves layers unchanged. 



\begin{figure*}[t]
    \centering
    \begin{minipage}[b]{1\linewidth}
    \centering
    \includegraphics[width=1\textwidth]{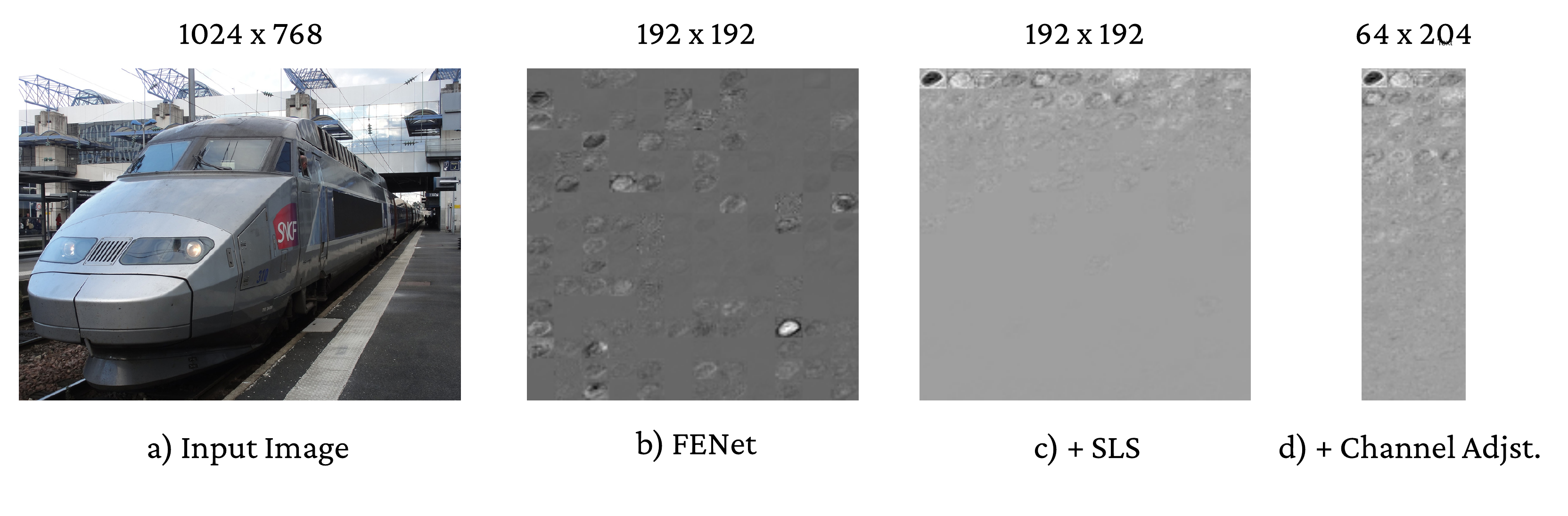}
    \end{minipage}
\caption{Examples of the feature reduction and conversion process with different tool selection. a) The original image before NN-Part 1. b) FENet turned on. c) FENet + SLS turned on. d) FENet + SLS + Channel Adjustment turned on. The addition of each tool improves compressibility with the feature inner encoder.}
\vspace{-0.5cm}
\label{fig:reduced_features}
\end{figure*}

\subsection{Transform}
The Transform module is responsible for reducing the volume of intermediate features that need to be transmitted to the decoder. In typical backbone architectures such as ResNet with FPN~\cite{faster_rcnn}, widely used in frameworks like Detectron2~\cite{detectron2}, the feature extractor outputs four multi-scale feature layers—P2, P3, P4, and P5—after the designated split point. These feature layers correspond to progressively coarser spatial resolutions, where P2 has the highest and P5 the lowest resolution. Despite differing spatial sizes between layers, each feature map typically contains 256 channels, resulting in a total of 1024 channels. This combined representation can require significantly more bandwidth than the original input image in terms of total elements.

To solve this issue, FCM employed a learned network module i.e.,  Feature Fusion and Encoding Network, or FENet~\cite{feature_reduction} within the FCM pipeline. FENet takes several feature layers as input. The number of these input layers determines how many ``Encoding Blocks'' FENet uses. Each Encoding Block's specifics, including its 2-D convolutions, Residual Blocks, and Attention Blocks, depend on a predefined neural network split point. The first feature layer goes through the initial Encoding Block, which halves its spatial resolution. This result is then combined with the second feature layer before being processed by the second Encoding Block. This continues until the last feature layer has been processed. Finally, the resulting features are adjusted by a ``gain vector'' for each channel. This vector either boosts or diminishes channels based on how important they are learned to be, offering a balance between rate and distortion. The final output of FENet is a set of ``reduced features,'' which have half the spatial resolution of the smallest input feature layer, fewer total channels, and less important channels scaled closer to zero. This process effectively concentrates semantic information into a more compact representation, enabling efficient transmission with minimal task accuracy degradation.

The Selective Learning Strategy (SLS)~\cite{m71202} is an additional tool for controlled channel organization. Figure~\ref{fig:reduced_features}(b) and Figure~\ref{fig:reduced_features}(c) demonstrate the effect of without and with SLS respectively. Without SLS, the location of important channels is randomly distributed. With SLS, channels are ordered so lower indexed are more important than higher indexed. The later channels contain less structured information and are simpler to compress with the inner codec.

\subsection{Channel Adjustment}

To improve compression efficiency, Channel Adjustment performs channel-wise truncation on reduced feature tensors by identifying and removing channels with minimal activation~\cite{m68965}. These low-activation channels often carry little task-relevant information but still consume bits during transmission, potentially degrading overall system performance. Figure~\ref{fig:reduced_features}(d) shows the application of Channel Adjustment, where channel count has been significantly reduced.

The selection process relies on analyzing the dynamic range of each channel in the tensor. For each channel \( z_c \), the range \( r_c \) is calculated as the difference between its maximum and minimum values across spatial dimensions:
\begin{align}
r_c = \max(z_c) - \min(z_c)
\end{align}
This range provides a simple yet effective proxy for activation strength. Channels with higher ranges are likely to carry stronger, more varied signal content, while channels with lower ranges may be close to constant and thus considered inactive.

To distinguish active from inactive channels, a threshold \( T \) is computed by scaling down the mean of all channel ranges, ensuring that \( T \) lies below the average activation level across channels. This data-driven threshold requires no manual tuning and adapts to the statistical distribution of each frame.

Based on this threshold, a binary activity map is constructed as:
\begin{align}
a_c =
\begin{cases}
0, & \text{if } r_c \geq T \\
1, & \text{if } r_c < T
\end{cases}
\quad \text{for } c = 1, 2, \dots, C
\end{align}
where \( C \) is the total number of channels in \(z\), and \( a_c \in \{0, 1\} \) indicates whether channel \( c \) is removed (1) or maintained (0). This information is sent once per intra period.


\section{FEATURE CONVERSION}

The Feature Conversion module contains the following three tools: 1) Calculate Statistical Parameters, 2) Packing, and 3) Quantization.

\subsection{Calculate Statistical Parameters}
This module aims to align the distribution of the restored reduced feature with that of the original reduced feature~\cite{eimran_iscas}, denoted as $z$. Assuming $z$ follows a normal distribution, a single set of statistical parameters is derived and transmitted to the decoder~\cite{ce42b}. These parameters are encoded using 32-bit floating point precision and signaled every refinement period, aligned with the intra period of the inner codec. On the encoder side, the mean $\mu_z$ and standard deviation $\sigma_z$ of $z$ are computed as:

\begin{align}
\mu_z &= \frac{1}{\|z\|} \sum_{i}^{\|z\|} z[i] \\
\sigma_z &= \sqrt{\frac{1}{\|z\|} \sum_{i}^{\|z\|} (z[i] - \mu_z)^2}
\end{align}

\subsection{Packing}
After the feature reduction stage, the resulting array of 2-D tensors are spatially reorganized into a single-channel 2-D frame using a raster scan layout. To enable accurate reconstruction, the original dimensions are transmitted as side information to the decoder.

\subsection{Quantization}
A min-max normalization technique~\cite{m67513} is applied to scale the values of the input tensor $z$ within a fixed range of $[0, 1]$. This is achieved by computing the minimum and maximum values of $z$, denoted as $z_{\min}$ and $z_{\max}$, and applying the following transformation:

\begin{equation}
z_{norm} = \min\left( \max\left( \frac{z - z_{\min}}{z_{\max} - z_{\min}}, 0 \right), 1 \right)
\end{equation}

\noindent where $z_{norm}$ represents the normalized output.

After normalization, uniform quantization is applied to map the normalized values to discrete levels. Given a bit depth of $\textit{bitdepth}$, the maximum number of quantization levels is computed as $\textit{max\_num\_bits} = 2^{\textit{bitdepth}}$. The quantized output $z_q$ is then obtained by:

\begin{multline}
z_q = \min\Bigg( \max\Bigg( 
\left\lfloor z_{norm} \times \textit{max\_num\_bits} \right\rfloor, 
0 \Bigg),\\ \textit{max\_num\_bits} - 1 \Bigg)
\end{multline}

\noindent where $\lfloor \cdot \rfloor$ denotes the floor operation. This operation ensures that the quantized values are clipped within the valid range of $[0, 2^{\textit{bitdepth}} - 1]$.

\section{FEATURE ENCODING}
The packed and quantized feature frame, formatted as 10-bit monochrome YUV (4:0:0), is encoded using a conventional image/video codec such as Versatile Video Coding (VVC)~\cite{vvc}, which provides high spatial and temporal compression efficiency. The encoder is operated in low-delay mode to reflect real-time streaming conditions.

\section{FEATURE DECODING}
The decoding process utilizes the image/video decoder such as the VVC decoder~\cite{vvc} to reconstruct the 10-bit monochrome frames from the compressed bitstream.

\section{FEATURE INVERSE CONVERSION}

The Inverse Feature Conversion module contains the following three tools: 1) Dequantization, 2) Unpacking, and 3) Reduced Feature Refinement.

\subsection{Dequantization}
A linear dequantization strategy is employed to recover normalized feature values from the quantized representation~\cite{ce42b}. Given a quantized feature tensor $z_q$ and a predefined bit depth $\textit{bitdepth}$, the maximum quantization level is computed as $\textit{max\_num\_bits} = 2^{\textit{bitdepth}} - 1$. The dequantized feature tensor $z_{dq}$ is then obtained by applying the following linear scaling:

\begin{equation}
z_{dq} = \frac{z_q}{\textit{max\_num\_bits}}
\end{equation}

\noindent This operation restores the feature values to the normalized range $[0, 1]$.

\subsection{Unpacking}
Unpacking reverses the spatial transformation applied during the packing step. The decoded frame is reshaped into its original multi-channel configuration. The dimensions used for reshaping are recovered from the transmitted side information.

\subsection{Reduced Feature Refinement}
Reduced feature distribution refinement is defined as follows:

\begin{equation}
z = \left( \frac{z_{dq} - \mu_{z_{dq}}}{\sigma_{z_{dq}}} \right) \cdot \sigma_z + \mu_z
\end{equation}

\noindent where $\mu_{z_{dq}}$ and $\sigma_{z_{dq}}$ denote the mean and standard deviation of the unrefined reduced feature $z_{dq}$, respectively. The parameters $\mu_z$ and $\sigma_z$ represent the statistical properties of the original reduced feature, provided by the encoder in 32-bit floating-point format, and are updated periodically according to a predefined refinement refresh interval~\cite{eimran_iscas}.

\section{FEATURE RESTORATION}

The Feature Restoration module contains the following four tools: 1) Inverse Channel Adjustment, 2) Inverse Transform, 3) Temporal Upsampling, and 4) Restored Feature Refinement.

\subsection{Inverse Channel Adjustment}

The binary vector \( \mathbf{a} = [a_1, a_2, \dots, a_C] \), transmitted as side information, indicates which channels were removed (\( a_c = 1 \)) or kept (\( a_c = 0 \)) during encoding. Its length \( C \) matches the original number of channels, allowing the decoder to restore the full tensor shape.

At the decoder, the transmitted channels are inserted into their original positions based on \( \mathbf{a} \). Removed channels are reconstructed by filling them with the average of the decoded active (\( a_c = 0 \)) channels. This preserves original tensor dimensions.

\subsection{Inverse Transform}

The inverse transform takes the compressed features and reconstructs the original intermediate features using a neural network called DRNet~\cite{feature_reduction}. DRNet first reverses the scaling applied during compression and then progressively restores the spatial and channel-wise details through a series of decoding stages. Each stage employs standard neural network operations such as convolution and attention to enhance the feature representation. At every step, the output is split: one branch adjusts the channels to improve reconstruction quality, while the other propagates to the next decoding stage. This process continues iteratively until all original feature layers are fully recovered.

\subsection{Temporal Upsampling}

If temporal resampling is enabled, missing sets of feature layers need to be reconstructed in order to preserve the order received from NN-Part 1 to the encoder~\cite{m69884}. As the sampling ratio is $2\times$, a discarded set of layers will have one past set and one future set that immediately neighbor it in the temporal order. These do get preserved by the encoder and serve as references for reconstruction. Linear interpolation is performed between the two references to produce a new set of feature layers. Specifically, with reference layers \(y^{past}_n\) and \(y^{future}_n\), feature layer \(y_n\) is determined by Equation (\ref{eq:temporal_upsampling}).

\begin{equation}
y_n=\frac{y^{past}_n + y^{future}_n}{2}
\label{eq:temporal_upsampling}
\end{equation}

\subsection{Restored Feature Refinement}
The distribution of the unrefined restored intermediate feature tensor $Y$ is adjusted using quantized statistical parameters: the mean $\mu_{X}$ and standard deviation $\sigma_{X}$, both of which are transmitted by the encoder. This refinement yields the statistically refined tensor $\hat{X} = \{\hat{x}_1, \hat{x}_2, \dots, \hat{x}_N\}$.

The distribution of the $n$-th refined intermediate feature tensor $\hat{x}_n$ is computed as follows:

\begin{equation}
\hat{x}_n = \left( \frac{y_n - \sum_{m=1}^{N} \mu(y_m)}{\sqrt{\sum_{m=1}^{N} \sigma^2(y_m)}} \right) \cdot \sigma_{X} + \mu_{X}
\end{equation}

\noindent where $y_n$ denotes the $n$-th unrefined intermediate feature tensor within $Y$. This refinement ensures that the statistical distribution of the decoded features aligns more closely with the encoder-side distribution~\cite{eimran_iscas}.
\section{EXPERIMENTAL RESULTS}
\begin{table*}[t]
\normalsize
\begin{center}
\caption{Common Test \& Training Conditions (CTTC)}
\label{tbl:cttc}
\resizebox{0.95\linewidth}{!}{%
\begin{tabular}{llll}
\toprule
\textbf{Dataset} & \textbf{Task} & \textbf{Network} & \textbf{Split Points} \\ \midrule
OpenImagesV6~\cite{oiv6_seg} & Instance segmentation & MaskRCNN-X101-FPN~\cite{mask_rcnn} & $\{ p_2, p_3, p_4, p_5 \}$ \\
OpenImagesV6~\cite{oiv6_det} & Object detection       & FasterRCNN-X101-FPN~\cite{faster_rcnn} & $\{ p_2, p_3, p_4, p_5 \}$ \\
SFU~\cite{sfu_v1}            & Object detection       & FasterRCNN-X101-FPN~\cite{faster_rcnn} & $\{ p_2, p_3, p_4, p_5 \}$ \\
TVD~\cite{tvd}               & Object tracking        & JDE-1088x608~\cite{tracking}           & $\{ d_{36}, d_{61}, d_{74} \}$ \\
HiEve~\cite{hieve}           & Object tracking        & JDE-1088x608~\cite{tracking}           & $\{ d_{105}, d_{90}, d_{75} \}$ \\
\bottomrule
\end{tabular}
}
\end{center}
\end{table*}
\label{sec:exp_results}
In this section, we describe the experimental setup, performance, and complexity analysis of FCM. For our experiments, we have used Feature Coding Test Model (FCTM) version 6.1.
\footnote{FCTM Version 6.1: \url{https://git.mpeg.expert/MPEG/Video/fcm/fctm/}}

\begin{table*}[t]
\begin{center}
\normalsize
\caption{BD-Rate and Complexity Analysis of FCM Compared to Remote Inference}
\label{tbl:bd_rate_complexity}
\resizebox{1\linewidth}{!}{%
\begin{tabular}{@{}llc>{\centering\arraybackslash}m{3.5cm}>{\centering\arraybackslash}m{3.5cm}@{}}
\toprule
\textbf{Task Network} & \textbf{Dataset} & \shortstack{\textbf{BD-Rate}} & \shortstack{\textbf{Encoding} \\ \textbf{Complexity Ratio}} & \shortstack{\textbf{Decoding} \\ \textbf{Complexity Ratio}} \\ 
\midrule
Instance Segmentation & OpenImagesV6     & -94.24\% & 6.15  & 0.26 \\
\midrule
Object Detection      & OpenImagesV6     & -95.45\% & 14.34 & 0.22 \\
                      & SFU (Class A/B)  & -38.13\% & 3.02  & 0.28 \\
                      & SFU (Class C)    & -85.55\% & 3.41  & 0.48 \\
                      & SFU (Class D)    & -85.91\% & 2.59  & 0.36 \\
\midrule
Object Tracking       & TVD              & -94.57\% & 0.78  & 0.19 \\
                      & HiEve (1080p)    & -94.58\% & 0.43  & 0.12 \\
                      & HiEve (720p)     & -92.67\% & 0.43  & 0.12 \\
\midrule
\textbf{Overall}       &                  & \textbf{-85.14\%} & \textbf{4.39} & \textbf{0.27} \\
\bottomrule
\end{tabular}
}
\end{center}
\end{table*}

\subsection{Experimental Setup}
We adopt the Common Test and Training Conditions (CTTC)~\cite{fcm_cttc} defined by MPEG to generate and compare the performance results of FCM. These common test conditions are defined by industry experts to establish standardized protocols for benchmarking coding-for-machine systems under real-world constraints. Table~\ref{tbl:cttc} summarizes the datasets, vision tasks, backbone network architectures, and split points used in our experiments by following the CTTC.

Our evaluation spans five benchmark datasets and covers three core computer vision tasks: instance segmentation, object detection, and object tracking. We employ three representative neural network architectures: Mask R-CNN with ResNeXt-101 backbone~\cite{mask_rcnn} for instance segmentation, Faster R-CNN with a ResNeXt-101 backbone~\cite{faster_rcnn} for object detection, and JDE~\cite{tracking} for object tracking. These networks are widely adopted and serve as strong baselines in the respective domains.

For image datasets, we have used the All-Intra~\cite{vvc} configuration of the Versatile Video Coding (VVC) Test Model version 23.0 (VTM 23.0)~\cite{vtm} for encoding and decoding, while the Low-Delay~\cite{vvc} configuration is employed for video data sets.

\subsection{BD-Rate Performance}
We assess the efficiency of the proposed FCM framework using the Bjøntegaard Delta Rate (BD-Rate) metric~\cite{bd_rate}, which measures the percentage bitrate savings at equivalent task accuracy. All evaluations are performed using the CompressAI-Vision toolkit~\cite{compressai_vision, m65437}.

Table~\ref{tbl:bd_rate_complexity} summarizes the BD-Rate performance achieved by FCM compared to conventional remote inference. In remote inference, images/videos are first compressed and sent to the remote server. Then in the remote server, inferences are done on the decompressed inputs. This is the most popular and conventional way for machine vision analysis as many edge devices cannot run large neural networks.

FCM achieves the same task accuracy while using, on average, 85.14\% fewer bits compared to the conventional remote inference method. Notably, it reduces bandwidth by up to 94.58\% in object tracking tasks. These results highlight FCM’s effectiveness in minimizing communication overhead without compromising performance.

\subsection{Complexity Analysis}
While compression efficiency is a key advantage of FCM, it is equally critical to evaluate the computational overhead introduced by the encoding and decoding stages. Specifically, we define the following two complexity ratios to assess the feasibility of FCM under resource constraints:

\begin{equation}
\frac{\text{FCM Encoder Complexity}}{\text{NN Part 2 Complexity}} < 1
\label{eq:encoder_complexity}
\end{equation}

\begin{equation}
\frac{\text{FCM Decoder Complexity}}{\text{NN Part 1 Complexity}} < 1
\label{eq:decoder_complexity}
\end{equation}

Equation (\ref{eq:encoder_complexity}) compares the computational cost of the encoder (run on the edge device) with the cost of the remaining portion of the neural network that would otherwise be executed locally. A value less than one indicates a net savings in edge-side computation. Equation (\ref{eq:decoder_complexity}) serves an analogous purpose on the server side.

Table~\ref{tbl:bd_rate_complexity} presents the measured complexity ratios for each task and dataset. All models were executed on CPU using an AMD EPYC 7702 64-Core processor, and wall-clock execution time was used as a proxy for complexity. 

Notably, while the decoder is lightweight in all cases (average ratio of 0.27), the encoder incurs significantly higher computational overhead (average ratio of 4.39), particularly in object detection tasks. This is primarily due to the complexity of the Feature Encoding Network (FENet) and Decoding and Restoration Network (DRNet). 

\section{CONCLUSION \& FUTURE DIRECTION}

Feature Coding for Machines (FCM) offers a transformative approach for scaling deep learning across distributed, resource-constrained platforms by enabling efficient split inference. Through standardized bitstream syntax and intelligent feature transformation, FCM compresses intermediate neural activations with minimal impact on task performance—achieving over 85\% average bitrate reduction across object detection, segmentation, and tracking tasks. These gains open new possibilities for AI-driven consumer devices where bandwidth, privacy, and energy are critical constraints.

While the current FCM implementation shows strong compression performance, it still incurs encoder-side overhead due to complex feature reduction and transformation networks. Its reliance on task-specific modules also limits out-of-the-box interoperability across diverse models. Looking ahead, continued research into lightweight, task-agnostic coding architectures is essential. Achieving universal compatibility without retraining—and reducing edge-side computation—will be key to unlocking FCM’s potential in real-time, low-power, and privacy-sensitive applications. As the standard evolves, FCM paves the way for scalable, interoperable machine vision in next-generation consumer electronics.




\newpage

\begin{IEEEbiography}{Md Eimran Hossain Eimon}{\,}
received his M.S. in Computer Science from Florida Atlantic University, Boca Raton, FL, USA, where he is currently pursuing a Ph.D. in Computer Science. His research interests include coding for machine vision and video compression. Contact him at meimon2021@fau.edu.
\end{IEEEbiography}

\begin{IEEEbiography}{Juan Merlos}{\,}received his M.S. in Computer Engineering from Florida Atlantic University, Boca Raton, FL, USA, where he is currently working toward a Ph.D. in Computer Science. His research interests include computer vision and data compression. Contact him at jmerlosjr2017@fau.edu.
\end{IEEEbiography}

\begin{IEEEbiography}{Ashan Perera}{\,} recieved his M.S. in Computer Science from Florida Atlantic University, Boca Raton, FL, USA. His research interests include computer vision and data compression. Contact him at aperera2016@fau.edu.
\end{IEEEbiography}

\begin{IEEEbiography}{Hari Kalva}{\,} is a Professor and Chair in the Department of Electrical Engineering and Computer Science at Florida Atlantic University. His research focuses on video coding and multimedia systems. He has made significant contributions to video codec standardization through active participation in MPEG, and he is also a senior member of IEEE with numerous publications and patents in multimedia technologies. Contact him at hkalva@fau.edu.
\end{IEEEbiography}

\begin{IEEEbiography}{Velibor Adzic} {\,} received a Ph.D in Computer Science from Florida Atlantic University, Boca Raton, FL, USA, in 2014, where he is currently an Assistant Professor of Teaching. His interests include video and image compression. Contact him at vadzic@fau.edu.
\end{IEEEbiography}

\begin{IEEEbiography}{Borko Furht} {\,} is a professor in the Department of Electrical Engineering and Computer Science at Florida Atlantic University, Boca Raton, FL, USA. His research interests include multimedia systems, video coding and compression, 3D imaging, big data analytics, and cloud computing. He is the founding Editor-in-Chief of Multimedia Tools and Applications and co-founder of the Journal of Big Data. Contact him at bfurht@fau.edu.

\end{IEEEbiography}

\end{document}